\documentclass[letterpaper, 10 pt, conference]{ieeeconf}  

\IEEEoverridecommandlockouts                              
                                                 
\usepackage{graphics} 
\usepackage{epsfig} 
\usepackage{mathptmx} 
\usepackage{times} 
\usepackage{amsmath} 
\usepackage{amssymb}  
\usepackage{url}
\usepackage{romannum}
\usepackage{xcolor}
\usepackage{float}
\usepackage{url}
\usepackage[ruled,vlined]{algorithm2e}
\usepackage{array}
\usepackage{placeins}
\usepackage{adjustbox}
\title{\LARGE \bf
Data Efficient Visual Place Recognition Using Extremely JPEG-Compressed Images
}

\author{Mihnea-Alexandru Tomiță$^{1}$, Bruno Ferrarini$^{1}$, Michael Milford$^{2}$, Klaus McDonald-Maier$^{1}$, Shoaib Ehsan$^{1,3}$
\thanks{$^{1}$Authors are with the School of Computer Science and Electronic Engineering, University of Essex, CO4 3SQ, United Kingdom
        {\tt\small matomi@essex.ac.uk, bferra@essex.ac.uk, kdm@essex.ac.uk, sehsan@essex.ac.uk}. }
\thanks{$^{2}$Michael Milford is with the School of Electrical Engineering and Computer Science, Queensland University of Technology, Brisbane, QLD 4000, Australia
        {\tt\small michael.milford@qut.edu.au}}
\thanks{$^{3}$Shoaib Ehsan is also with the School of Electronics and Computer Science, University of Southampton, SO17 1BJ, United Kingdom
        {\tt\small s.ehsan@soton.ac.uk}}
\thanks{This work is supported by the UK Engineering and Physical Sciences Research Council through grants EP/R02572X/1 and EP/P017487/1.}
}

\begin{document}

     \maketitle
    \thispagestyle{empty}
    \pagestyle{empty}

     \begin{abstract}
     Visual Place Recognition (VPR) is the ability of a robotic platform to correctly interpret visual stimuli from its on-board cameras in order to determine whether it is currently located in a previously visited place, despite different viewpoint, illumination and appearance changes. JPEG is a widely used image compression standard that is capable of significantly reducing the size of an image at the cost of image clarity. For applications where several robotic platforms are simultaneously deployed, the visual data gathered must be transmitted remotely between each robot. Hence, JPEG compression can be employed to drastically reduce the amount of data transmitted over a communication channel, as working with limited bandwidth for VPR can be proven to be a challenging task. However, the effects of JPEG compression on the performance of current VPR techniques have not been previously studied. For this reason, this paper presents an in-depth study of JPEG compression in VPR related scenarios. We use a selection of well-established VPR techniques on well-established benchmark datasets with various amounts of compression applied. We show that by introducing compression, the VPR performance is drastically reduced, especially in the higher spectrum of compression. Moreover, this paper demonstrates how fine-tuning a CNN can be utilised as an optimisation method for JPEG compressed data to perform more consistently with the image transformations detected in extremely JPEG compressed images.

     
      \end{abstract}

      
      \section{Introduction}\label{introduction}

    Several robotic applications benefit from deploying multiple units operating in parallel, such as search and rescue missions, where multiple robots can cover the search area easier than a single agent. This is also the case for planetary exploration, where the dangerous terrain found on other planets can be simultaneously explored by multiple smaller robots. Moreover, due to the extreme nature of the task, employing multiple agents may be desirable to minimise the risk of failure when areas are too dangerous for only a single robot to explore. Multi-agent collaborative tasks are also of great interest to NASA, as deep space exploration continues. In \cite{schoolcraft2017marco}, two MarCO spacecrafts have been utilised to simultaneously explore the terrain of Mars, while relaying the captured data back to Earth. However, when a large number of robots are part of a decentralised system, the bandwidth must accommodate the entire system to facilitate swift data transmission between them. Moreover, when operating in bandwidth constrained environments, the robots are still required to transmit data \cite{nettleton2006decentralised}. 
    Knowing their position in the operating environment as well as the positions of the other agents is fundamental for mobile robots. As part of the visual simultaneous localisation and mapping (SLAM), visual place recognition (VPR) is an essential task for the localisation process when the environment is unstructured, GPS is unavailable or the visual odometry drifts due to accumulated errors. For applications that involve a single robot, an ideal VPR method is accurate in detecting known places and efficient enough to fit the hardware and battery capability of the deputed robot \cite{8792942}. For applications requiring multiple robotic platforms to collaborate, navigate and map the environment effectively, the visual data gathered must be transmitted remotely between each robot \cite{cieslewski2017efficient, cieslewski2018data, burguera2020unsupervised}.  Hence, the amount of data required to be transmitted must be taken into consideration when working with limited bandwidth available for VPR.
    
      \begin{figure}[t]
            \centering
            \begin{tabular}{ c }
                \includegraphics[width=234pt]{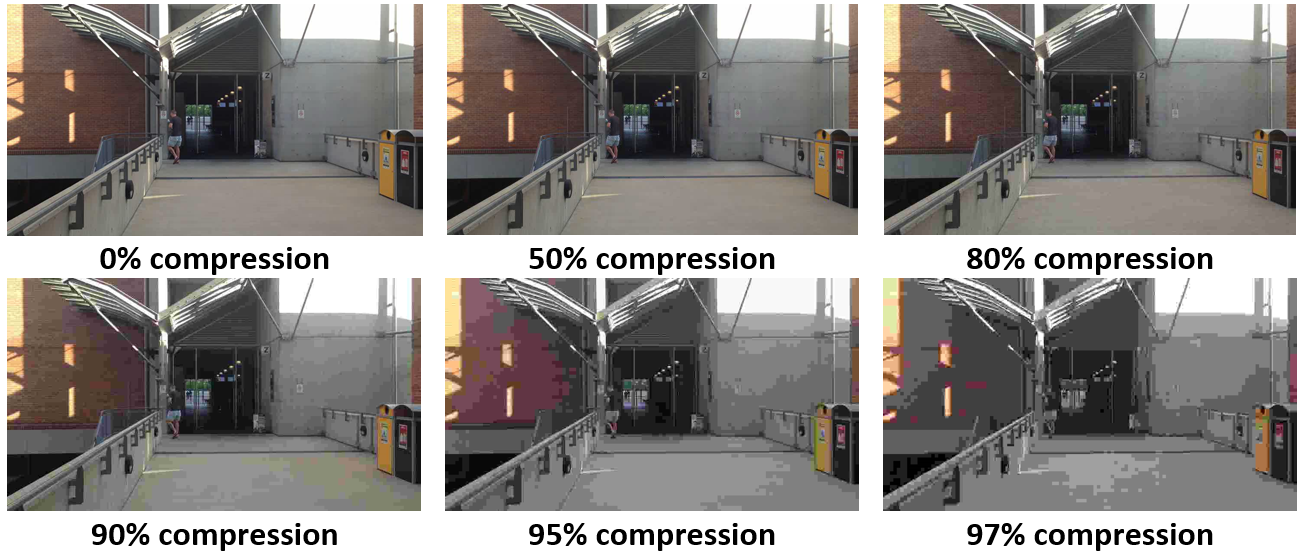}
            \end{tabular}
            \caption{The same image (taken from the Gardens Point \textit{day left} dataset) with different compression percentages (0\%, 50\%, 80\%, 90\%, 95\% and 97\%). It can be seen that above 90\% compression the image is drastically changed from the original uncompressed image.}
            \label{compression_effect}
            \end{figure}
    
    This paper proposes to use highly compressed JPEG images to reduce the amount of data that is transmitted in decentralized VPR contexts. 
    %
    %
    ISO/IEC-ITU JPEG is one of the most widely used compression standards employed to facilitate significant data storage and transmission reduction \cite{hudson_jpeg-1_2018}. In our opinion, the high compression ratios enabled by JPEG compared to other standard techniques \cite{8301939,mateika2007analysis} make it attractive for distributed VPR applications. However, JPEG is designed to have a minimal impact on the human perception system \cite{haines1992effects}. It is uncertain how the performance of various VPR techniques is affected throughout the compression spectrum. To the best of our knowledge, the study of JPEG for VPR application has been overlooked so far. To bridge this gap, this paper proposes to tune CNNs that deal with highly compressed JPEG images to circumvent the limitation of existing techniques. In summary, our contributions are as follows:

      
      
      
      

      \begin{itemize}
        \item An assessment of several well-established VPR techniques under mild to extreme JPEG compression rate, as shown in Fig. \ref{compression_effect}. This analysis uses several datasets presenting illumination, viewpoint, and weather variations to cover some of the most common viewing conditions experienced by a robot in real-world decentralized applications, where the operating environment might present heterogeneous conditions in different places (see Fig. \ref{datasetimagescompression}).
        \item We demonstrate how a fine-tuned CNN-based descriptor on highly JPEG compressed data can achieve higher and more consistent VPR performance than non-optimized VPR techniques. 
      \end{itemize}
      
      
      The remainder of this paper is organised as follows: Section \ref{literature_review} presents a literature review. In section \ref{methodology}, JPEG compression is presented, together with the place matching approach employed and the performance metric utilised in conducting our analysis. Section \ref{experimental_study} describes the utilised datasets and VPR techniques for performing the analysis on the effects of JPEG compression in VPR. Section \ref{results} presents the detailed results and analysis. Finally, the conclusions are presented in Section \ref{conclusion}.

            

      \section{Literature Review}   \label{literature_review}
    Environmental changes such as illumination \cite{sunderhauf2015performance} and viewpoint variation \cite{ESSEX3IN1} make a place appear differently on different traverses. These appearance changes render visual place recognition a challenging problem motivating significant effort put by the research community in proposing improvements to existing VPR methods and new techniques.
      Local feature descriptors such as Speeded-Up Robust Features (SURF) \cite{SURF} and Scale-Invariant Feature Transform (SIFT) \cite{lowe2004distinctive,SIFT} have been widely used in VPR \cite{se2002mobile}, \cite{andreasson2004topological}, \cite{stumm2013probabilistic}, \cite{kovsecka2005global}, \cite{murillo2007surf}. FAB-MAP \cite{cummins2011appearance} is an appearance based place recognition system that represents visual data as words and uses SURF for feature detection. Odometry information is included in FAB-MAP by the authors of \cite{maddern2012cat} to create CAT-SLAM. Center Surround Extremas (CenSurE) \cite{agrawal2008censure} can perform real-time detection and matching of features taken from images and has been utilised by FrameSLAM in \cite{konolige2008frameslam}. 
      Bag-of-Words model (BoW) \cite{4270197} and Vector of Locally Aggregated Descriptors (VLAD) \cite{jegou2010aggregating} build an image descriptor of fixed length by aggregating local feature descriptors around centroids. BoW and VLAD can be used for VPR as shown in \cite{jegou2010aggregating} and \cite{8792942}, respectively.
      Histogram-of-Oriented-Gradients (HOG) \cite{dalal2005histograms} is a global descriptor used to represent gradient angles, whilst also indicating the gradient magnitude for all image pixels. HOG is computationally efficient and tolerant to appearance changes \cite{zaffar2021vpr}. 
      Zaffar \textit{et al.} proposed in \cite{zaffar2020cohog} a training-free VPR system based on HOG feature descriptor, dubbed CoHOG, that has a low encoding time and good tolerance to lateral shift. 
      As shown in several comparison works \cite{8968579,zaffar2019levelling}, VPR descriptors work well with some place changes while not with others.  \cite{SwitchHit} proposes a switching system based on complementary of several VPR techniques \cite{9459537} to choose the best possible descriptor to deal with a particular place or environmental condition.
      
      The applicability of deep-learning in VPR has been originally studied by Chen \textit{et al.} in \cite{chen2014convolutional} where the authors combined all 21 layers of the Overfeat network \cite{sermanet2013overfeat} trained on ImageNet 2012 dataset with both the spatial and sequential filter of SeqSLAM. In \cite{chen2017deep}, the authors trained two neural network architectures on the Specific PlacEs Dataset (SPED) \cite{SPED}, more specifically AmosNet and HybridNet. Arandjelović \textit{et al.} \cite{arandjelovic2016netvlad}
      introduced a new layer which is based on a generalised VLAD entitled NetVLAD. Khaliq \textit{et al.} \cite{khaliq2019camal} presents a light-weight CNN-based VPR system, with low memory and resource utilization, that is robust to viewpoint and environmental changes. The authors of CALC \cite{merrill2018lightweight} present a light-weight system that is robust to both viewpoint and illumination variations. Cross-Region-Bow \cite{chen2017only} is a VPR technique that searches for regions of interest (ROIs) and uses the convolutional layer to create the representation of these salient regions. Khaliq \textit{et al.} present RegionVLAD  \cite{khaliq2019holistic}, a light-weight CNN-based VPR technique that can filter confusing elements and is able to detect salient features from images. While RegionVLAD is based on the same approach as Cross-Region-BoW, it uses VLAD for feature pooling. In \cite{AlexNet}, the authors have used the AlexNet ConvNet \cite{krizhevsky2012imagenet} pre-trained on the ImageNet ILSVRC dataset \cite{russakovsky2015imagenet} for object recognition. The authors of \cite{9725251} and \cite{HDBS} use binary neural networks (BNNs) for VPR. These systems are less computationally demanding than other CNN-based VPR techniques, while achieving similar place matching performance as full-precision systems. However, to run the BNNs, dedicated hardware or an inference engine that enables an efficient computation of bitwise operations is required. Efficiency is the main goal of Arcanjo et al. \cite{9672749} too, who propose to address VPR by taking inspiration from the drosophila neural system to create a lightweight neural network. 
      \endgraf
      
      In recent years, considerable effort has been put in creating decentralised VPR architectures. In contrast with centralised architectures where each robot sends the map to a central server that performs the place matching computations, in decentralised architectures the visual data gathered from each robot has to be shared between each robotic platform \cite{7811208}. To achieve a robust decentralised system, the bandwidth limitations of the communication network need to be overcome \cite{nettleton2006decentralised}. 
      The authors of \cite{9844233} propose a decentralised system for multi-robot exploration based on thermal images and inertial measurements. The front-end of the pipeline handles the feature tracking and place recognition, whilst the back-end component reduces both the memory and computational cost by utilising a covariance-intersection fusion strategy. The communication pipeline employed is based on VLAD, resulting in reduced bandwidth usage. In \cite{9341227}, a method for Multi-Robot SLAM based on ranging sensors is presented, where the system can create consistent maps even in scenarios where loop closures cannot be detected. In \cite{cieslewski2017efficient}, the descriptor space of NetVLAD is clustered, and each smaller cluster is sent to a robot to perform efficient decentralised VPR. In \cite{cieslewski2018data}, a data-efficient decentralised visual SLAM system is presented, where the data association scales linearly with the number of robots present in the Multi-Robot SLAM system. The authors of \cite{burguera2020unsupervised} present a loop detection architecture for performing Multi-Robot underwater visual SLAM. A common observation in \cite{7811208}, \cite{cieslewski2017efficient}, \cite{cieslewski2018data} and \cite{burguera2020unsupervised} is that in Multi-Robot SLAM systems, working with a reduced bandwidth can significantly increase the difficulty in transferring the images between multiple autonomous vehicles. This work addresses the data transfer problem using JPEG compression to facilitate VPR applications where the available bandwidth is not capable of transmitting the visual data in an uncompressed form.

      \section{Methodology} \label{methodology}
      
       \subsection{Image Compression} \label{imagecompression}


      JPEG is a compression method that allows the user to select and adjust the amount of compression applied to an image. As JPEG is a lossy compression method, there is a trade-off between image clarity and image size. By applying an increased amount of compression to any given image (e.g. above 90\%), artifacts are introduced in the resulting compressed image, which could potentially lead to image alteration as seen in Fig. \ref{compression_effect}.

      The JPEG compression process can be broken down into three main steps. Firstly, the data in a given image is divided into the color and luminance components. As the human perception system is better suited to perceive intensity rather than color information, the latter can be subsampled to reduce the amount of data whilst maintaining the image visually unchanged to the user. Secondly, the data subsampled from the color component is divided into 8x8 pixel blocks. On each block, the Discrete Cosine Transform (DCT) is applied to describe the image content by the coefficients of the spatial frequencies for vertical and horizontal orientations, instead of pixel values. Finally, data quantization is performed, where the higher frequency coefficients are transformed to 0 first. Depending on the amount of compression selected by the user, the subsampling step may be skipped to achieve mild image compression. Conversely, to extremely compress an image, the subsampling step is turned on, and the quantization matrix is selected so that most coefficients are set to 0.
      
      The amount of compression applied to a given image is a parameter of the JPEG function \cite{hudson_jpeg-1_2018}, having values in range [0,99].
      As the visual quality of the image is not compromised in the lower spectrum of JPEG compression, a lower value (e.g. 50\%) should be selected to achieve mild image compression. Conversely, for an extremely JPEG compressed image, a high value (e.g. 97\%) should be assigned to the compression parameter.

      \subsection{Place matching}
      Place matching is performed by retrieving the best reference image from the map that corresponds with the visual data taken from a robot's camera, known as the query image. To find the best image in the map, each query image is matched with every reference image in the dataset, using the cosine \cite{chen2017only} to generate a similarity score between each query-reference pair:
     \begin{equation}
           s = \frac{Q_{F}  R_{F}}{||Q_{F}||\;||R_{F}||}\;\text{,}
            \label{eq:cosinesimilarity}
     \end{equation} 
     where \textit{$Q_{F}$} is the feature descriptor for a query image and \textit{$R_{F}$} is the feature descriptor for a reference image. For each query image, a distinct list \textit{S} of \textit{N} elements will be created containing the similarity coefficients as follows:
     \begin{equation}
           S = \{s_1,\,s_2,\, s_3,\, \dots{}\, s_N\}\;\text{,}
            \label{eq:similaritycoefficient}
     \end{equation}
      where \textit{N} is the total number of reference images in the dataset and \textit{s} is in range [0,1]. For each query image, the reference image with the highest similarity score is retrieved as the matching place.

      The matching process of a query image descriptor \textit{$Q_{F}$} with all the reference feature descriptors \textit{$R_{F}$} in the reference map \textit{$R_{M}$} can be summarized as follows. 
      In the first instance, the similarity scores are computed using equation (\ref{eq:cosinesimilarity}) between \textit{$Q_{F}$} and each \textit{$R_{F}$} in \textit{$R_{M}$}. Secondly, these scores are stored in a 1D array similar to equation (\ref{eq:similaritycoefficient}). Finally, the reference image that has the highest matching score (from $S$) is regarded as the matching place for \textit{$Q_{F}$}.
      
    \begin{table}[t]
            \caption{The size of each datasets in Megabytes (MB) with different compression percentages applied.} 
            \begin{center}
              \begin{tabular}{ |c|c|c|c|c|c|c| } 
            
            \hline
            & \multicolumn{6}{c|}{\textbf{Compression Applied}} \\\cline{2-7}
            \textbf{Dataset} & \textbf{0\%} & \textbf{50\%} & \textbf{80\%} & \textbf{90\%} & \textbf{95\%} & \textbf{97\%}\\
            \hline
            17 places & 81.8 & 18.9 & 10.8 & 8.3 & 6.6 & 5.9 \\
            \hline
            Nordland & 235.1 & 26 & 13.7 & 8.3 & 5.4 & 4.4 \\
            \hline
            Campus Loop & 46.8 & 9 & 4.6 & 2.6 & 1.5 & 1 \\
            \hline
            GP (day-to-day) & 54.8 & 17.6 & 8.4 & 5.2 & 3.1 & 2.3\\
            \hline
            GP (day-to-night) & 44.9 & 15.2 & 7 & 4.2 & 2.5 & 1.8\\
            \hline
            ORCD & 185.7 & 28.8 & 15.6 & 9.7 & 6.2 & 4.7 \\
            \hline
             ESSEX3IN1 & 1100 & 191.2 & 100 & 56.8 & 29.9 & 19.6 \\
            \hline
             SYNTHIA & 207.5 & 33.6 & 15.9 & 8.5 & 4.8 & 3.6\\
            \hline
            \end{tabular}
            \end{center}
            \label{table:image_size}
            \end{table}

       \subsection{Performance Metric}\label{employed_metrics}
       
       VPR performance is evaluated using the percentage of correctly matched images \cite{tomita2021convsequential,9849680}:
        
        \begin{equation}\label{eq:accuracy}
               Accuracy = \frac{N_c}{N_r}\;\text{,} 
        \end{equation} where $N_c$ represents the number of correctly matched query images and $N_r$ the total number of reference images in the map. The accuracy has values in range [0,1]. Higher the accuracy, higher the place matching performance of a VPR technique.
        
      \begin{figure*}[t]
            \centering
            \begin{tabular}{ c c }

                \includegraphics[width=240pt, trim=8 8 8 8, clip]{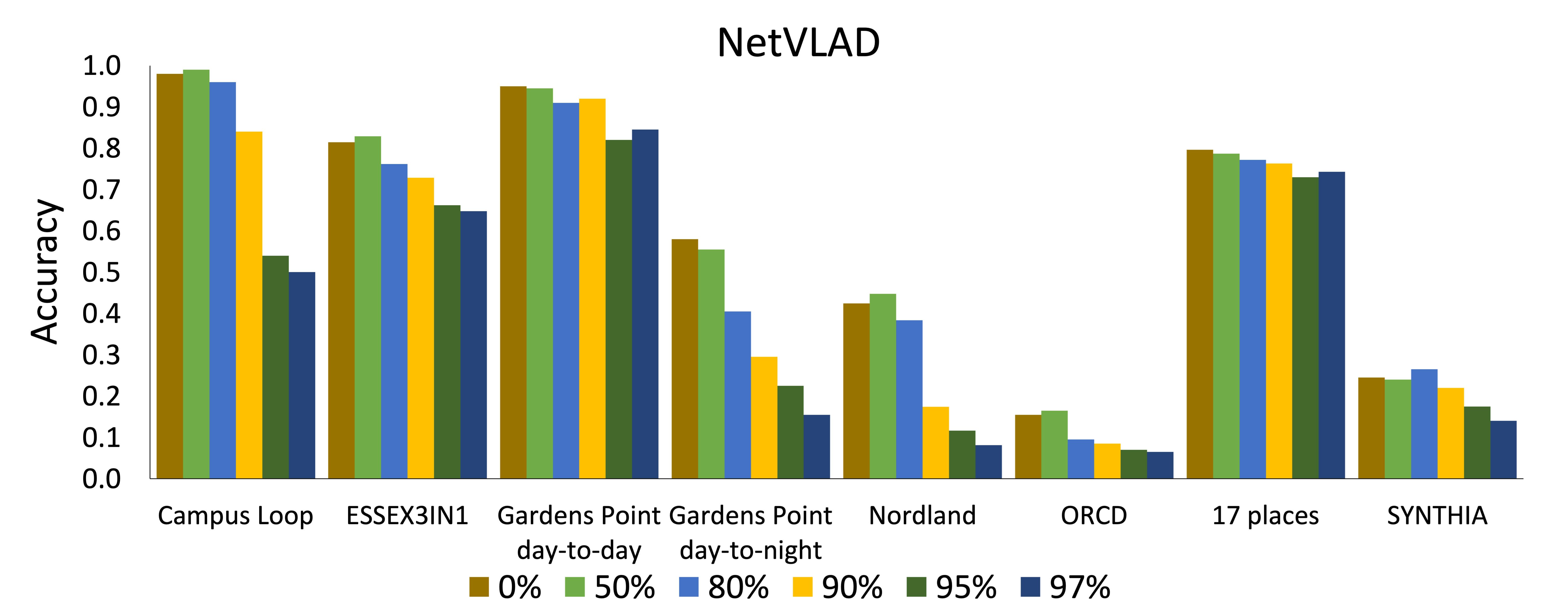} & 
                \includegraphics[width=240pt, trim=8 8 8 8, clip]{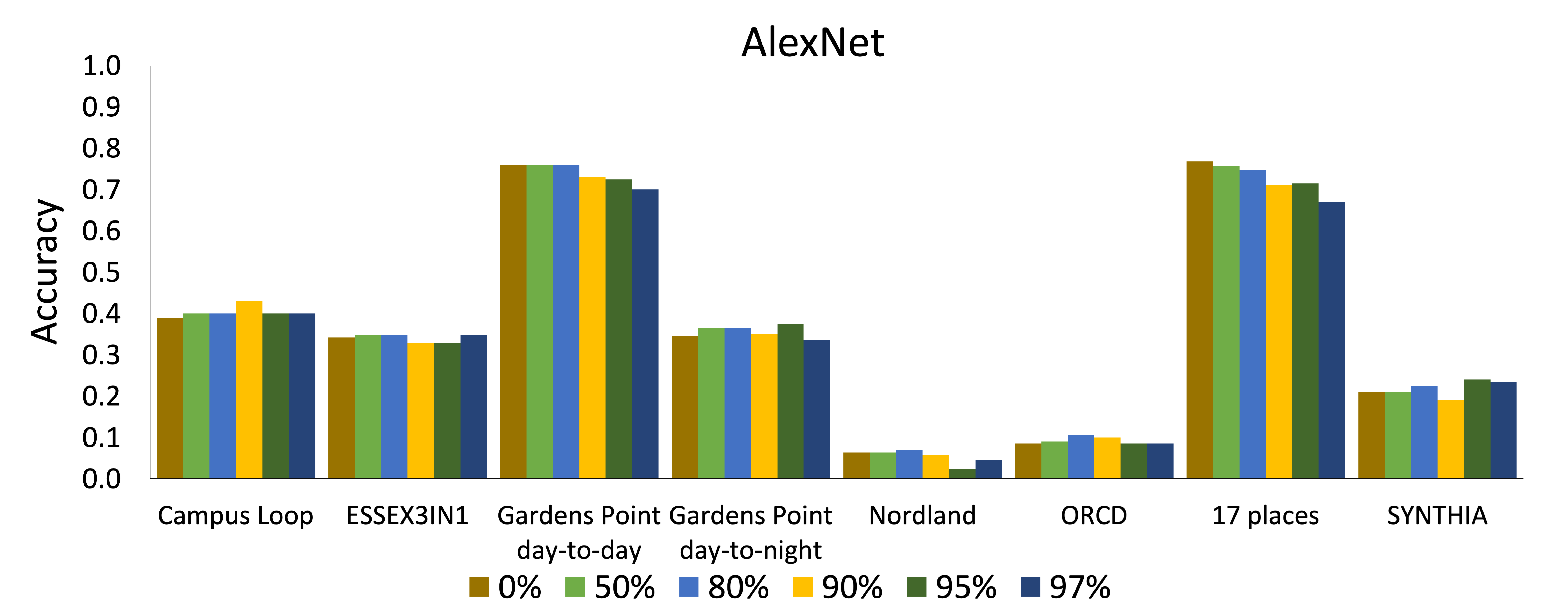} \\
                \includegraphics[width=240pt, trim=8 8 8 8, clip]{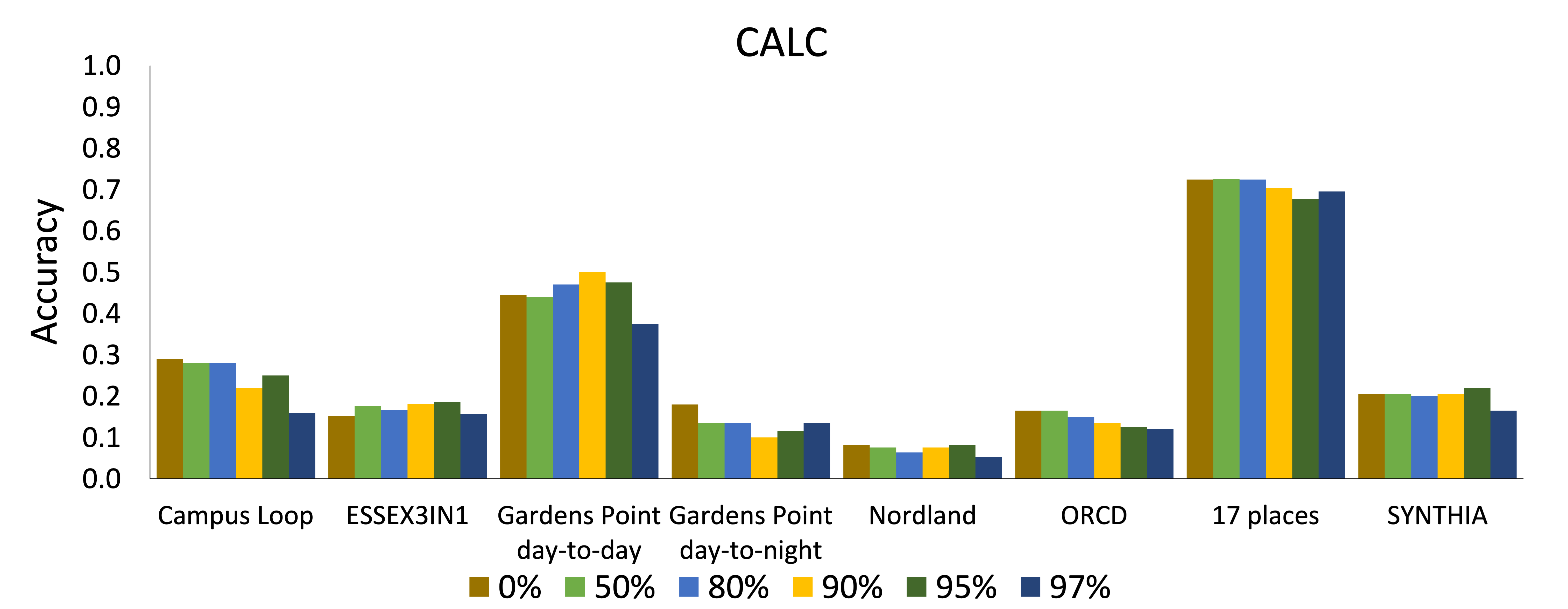}  &
                \includegraphics[width=240pt, trim=8 8 8 8, clip]{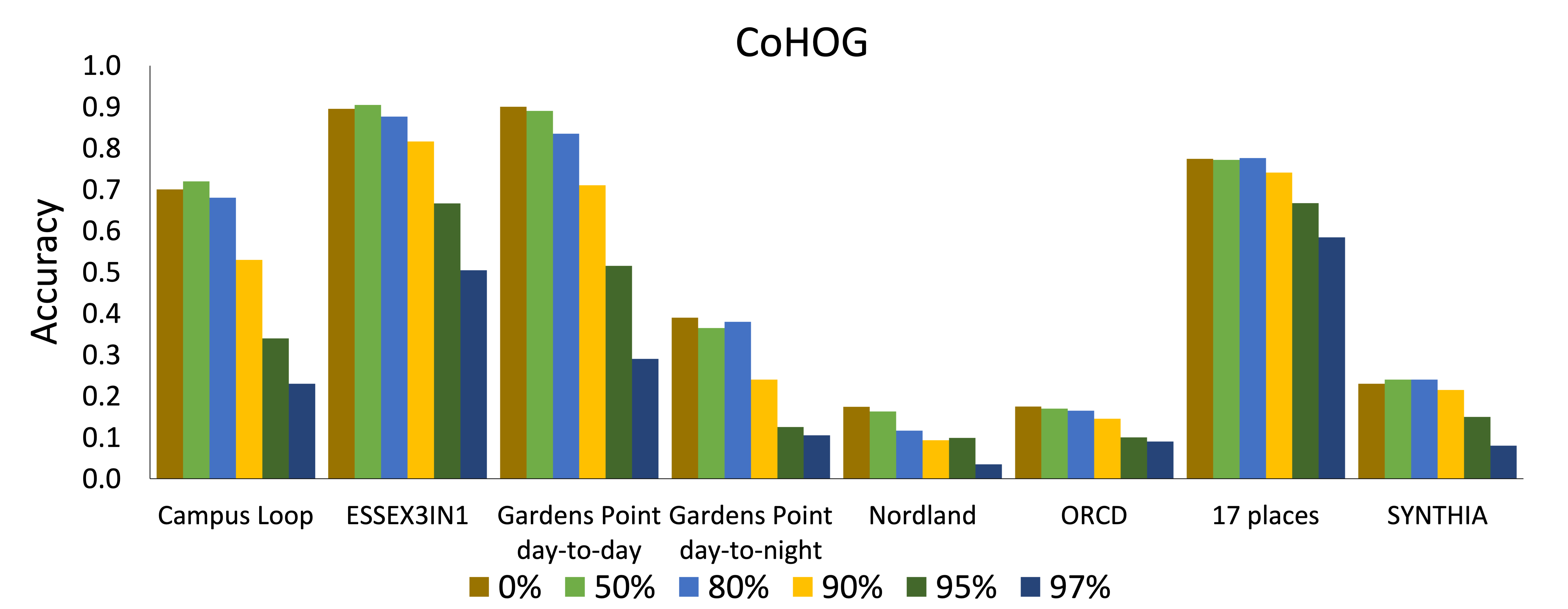}\\
                \includegraphics[width=240pt, trim=8 8 8 8, clip]{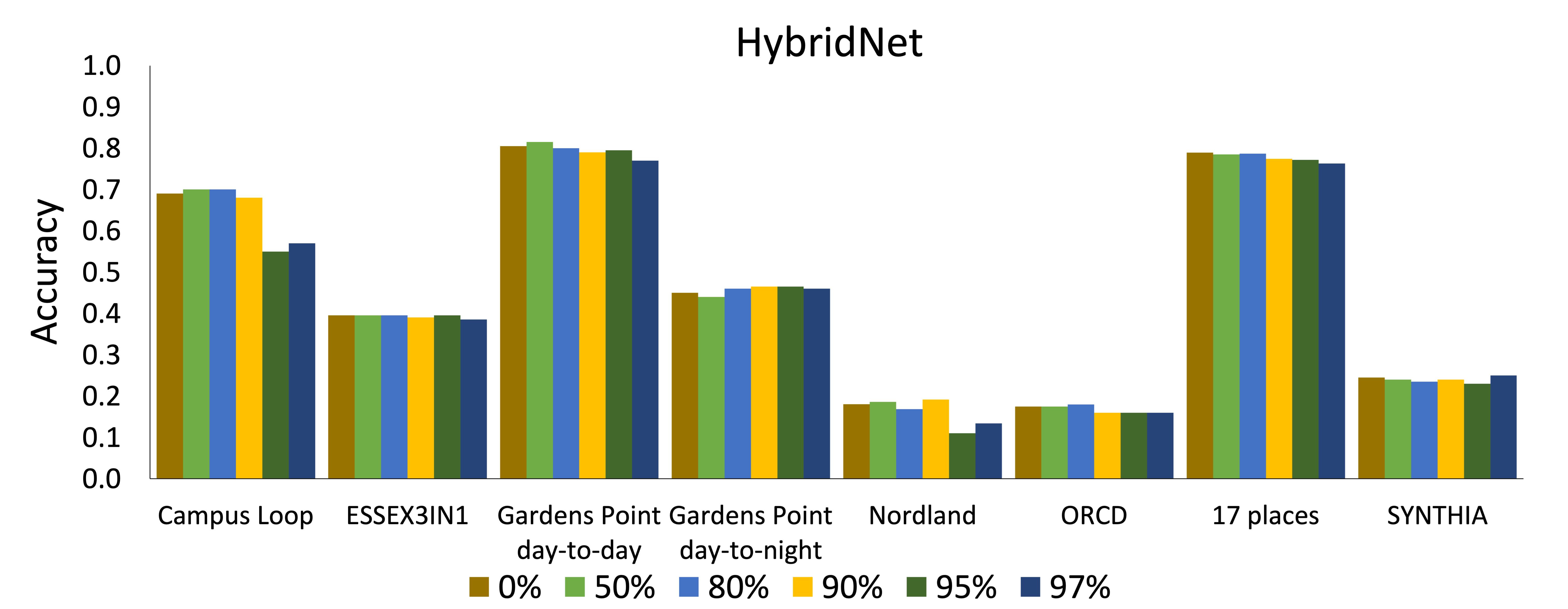}  &
                \includegraphics[width=240pt, trim=8 8 8 8, clip]{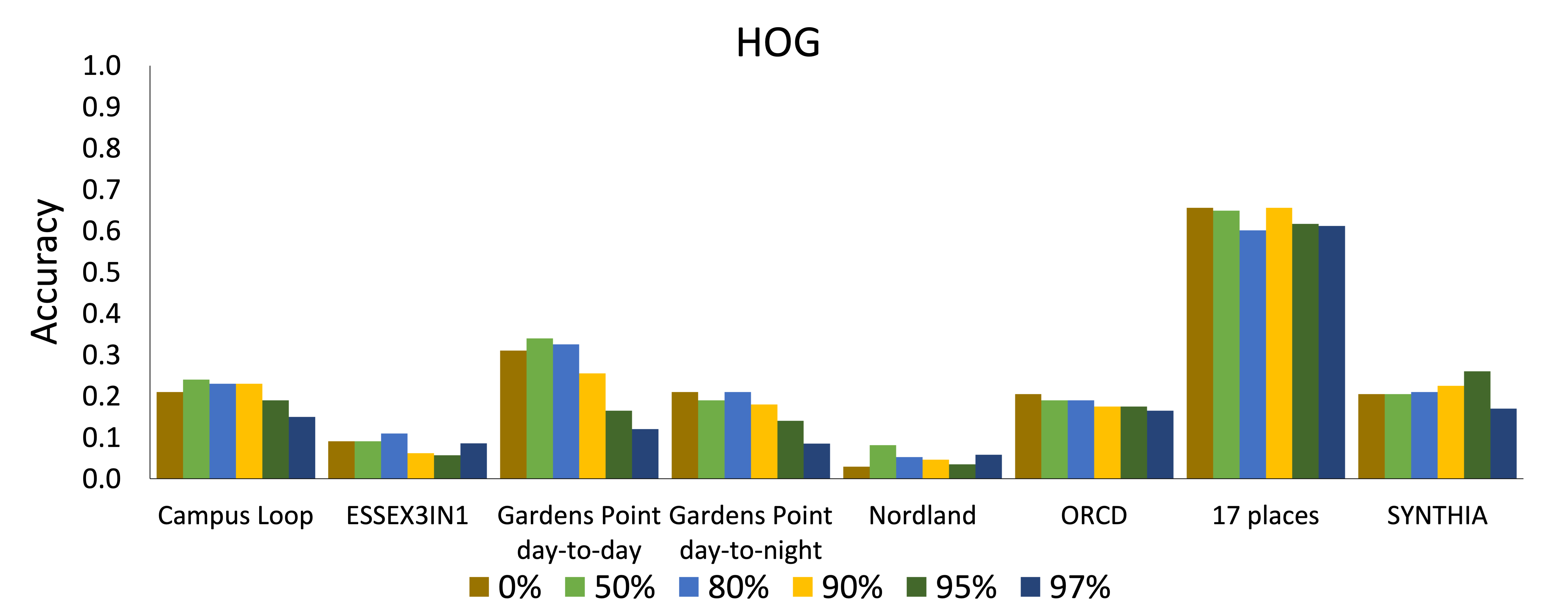}
                
            \end{tabular}
            \caption{The accuracy of all VPR techniques on each dataset with different levels of JPEG compression applied is presented here.}
            \label{VPRperformance}
            \end{figure*}
      
      \section{Experimental Setup} \label{experimental_study}
      This section presents the experimental setup for our work. We present the datasets used together with the VPR techniques employed for testing the effects of compression for VPR.
      
      \subsection{Test Datasets}\label{utiliseddatasets}
      The test data consists of eight datasets designed for VPR applications. Fig. \ref{datasetimagescompression} shows a query-reference pair for each of them. 
      The first dataset used is Campus Loop dataset \cite{merrill2018lightweight}. It contains 100 query and 100 reference images which pose challenges to any VPR system due to the high amount of viewpoint variation, seasonal variation and also the presence of statically-occluded frames. The second and third datasets are part of Gardens Point (GP) dataset \cite{sunderhauf2015performance} which contains both day and night images, that are divided as follows: 200 query images (\textit{day left}) and 400 reference images $-$ equally split into day images (\textit{day right}) and night images (\textit{night right}). Nordland dataset \cite{nordland2013dataset} is the fourth dataset used which captures the drastic visual changes that seasonal variation can have on a place (denoted by the changes in spring, summer, autumn and winter seasons). Since the most notable differences between seasons are seen during the summer and winter seasons, each VPR technique is tested on the summer-to-winter traverses of the Nordland dataset. The fifth dataset used is ESSEX3IN1 \cite{ESSEX3IN1}, which is composed of 420 images, equally split in 210 query and 210 reference images containing perceptual aliasing and confusing places/frames. 17 places \cite{sahdev2016indoor} is an indoor dataset, with images captured under illumination and viewpoint variation. Three locations have been selected namely Arena, AshRoom and Corridor that consist of 457 query (\textit{day\_vme1}) and 434 reference images (\textit{night\_vme1}).
      SYNTHIA dataset \cite{SYNTHIA} presents a simulated city-like environment, and contains frames in various weather, seasonal and illumination conditions. Oxford Robot Car dataset (ORCD) \cite{maddern20171} contains images under illumination and viewpoint changes. Table \ref{table:image_size} presents the size of each dataset (in Megabytes) for different levels of JPEG compression.
      

      \subsection{VPR Techniques}\label{utilisedvprtechniques}
      In this work, six well-established VPR techniques are used to show the effects of JPEG compression in VPR scenarios. These techniques are as follows: HOG \cite{dalal2005histograms}, CALC \cite{merrill2018lightweight}, HybridNet \cite{chen2017deep}, NetVLAD \cite{arandjelovic2016netvlad}, CoHOG \cite{zaffar2020cohog} and AlexNet \cite{AlexNet}. All VPR techniques are used as they are presented by their authors with no additional changes being made to neither technique. For a fair comparison with our model, the results for AlexNet have been generated utilising the \textit{fc6} layer.
      
      
      As mentioned in section \ref{imagecompression}, JPEG compression introduces artifacts while decreasing the quality of the image. As a result, JPEG compression introduces appearance changes in an image, rather than viewpoint changes. The selection of VPR techniques employed in this work can be divided into two main categories: VPR techniques that are robust to viewpoint changes (such as NetVLAD and CoHOG) and techniques that are optimized for appearance changes (such as HybridNet and AlexNet). Thus, we can identify which approach can be easily adapted to deal with extreme JPEG compression rates.
      


    \section{Results and Analysis}\label{results}
    In section \ref{placematchingperformance}, we present the effects of JPEG compression on the performance of several VPR techniques. We discuss the performance of each technique for several levels of compression in terms of accuracy. Furthermore, in section \ref{jpgoptimizedcnn}, the details of our JPEG optimized CNN are provided, whilst also presenting a comparison between our model trained on compressed data and other VPR techniques. In section \ref{nonuniformjpgcompression} we present the place matching performance of our model on non-uniform JPEG compressed data. 


    \subsection{Place Matching Performance}\label{placematchingperformance}
        
    By increasing the compression percentage on each dataset, we generally obtain lower results. This can be seen in Fig. \ref{VPRperformance}, where the accuracy (Y-axis) generally decreases with the increase in compression rate. This descending trend in performance is expected due to the fact that an increase in JPEG compression would conclude in a drastic change within the image structure (as observed in Fig. \ref{compression_effect}).

    The results presented in Fig. \ref{VPRperformance} show that the amount of compression applied to each dataset has a direct effect on the place matching performance. However, each technique is affected differently by image compression. A possible explanation for this decrease in place matching performance can be related to the inability of current VPR techniques to cope with the extreme changes that emerge from including image compression besides the already existing challenges in VPR (viewpoint, illumination, seasonal variations etc.). In particular, we observed that JPEG compression affects more those methods designed to deal with viewpoint changes such as NetVLAD and CoHOG. On the contrary, VPR descriptors that are designed to handle appearance changes present higher tolerance to JPEG compression. The details of our analysis are presented below.
     

    
    \begin{figure}[t]
            \centering
                 \begin{tabular}{ c }

                \includegraphics[width=234pt, trim=8 8 8 8, clip]{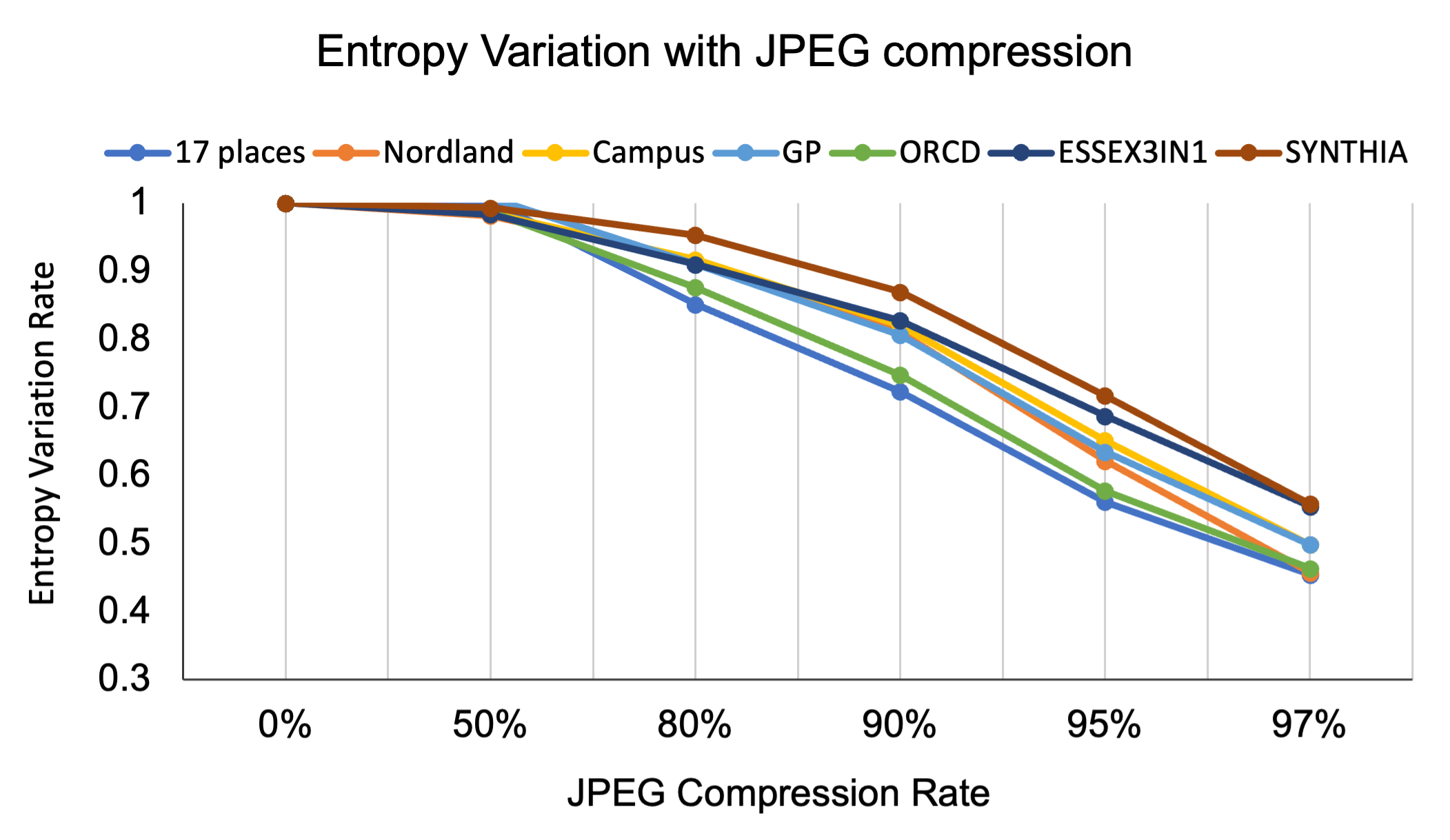}
                
            \end{tabular}
            \caption{Average entropy in query images with different compression ratios applied to each dataset.}
            \label{entropy_compression}
        \end{figure}
    
\begin{figure}[t]
            \centering
            \begin{tabular}{ c }
                \includegraphics[width=0.95\columnwidth]{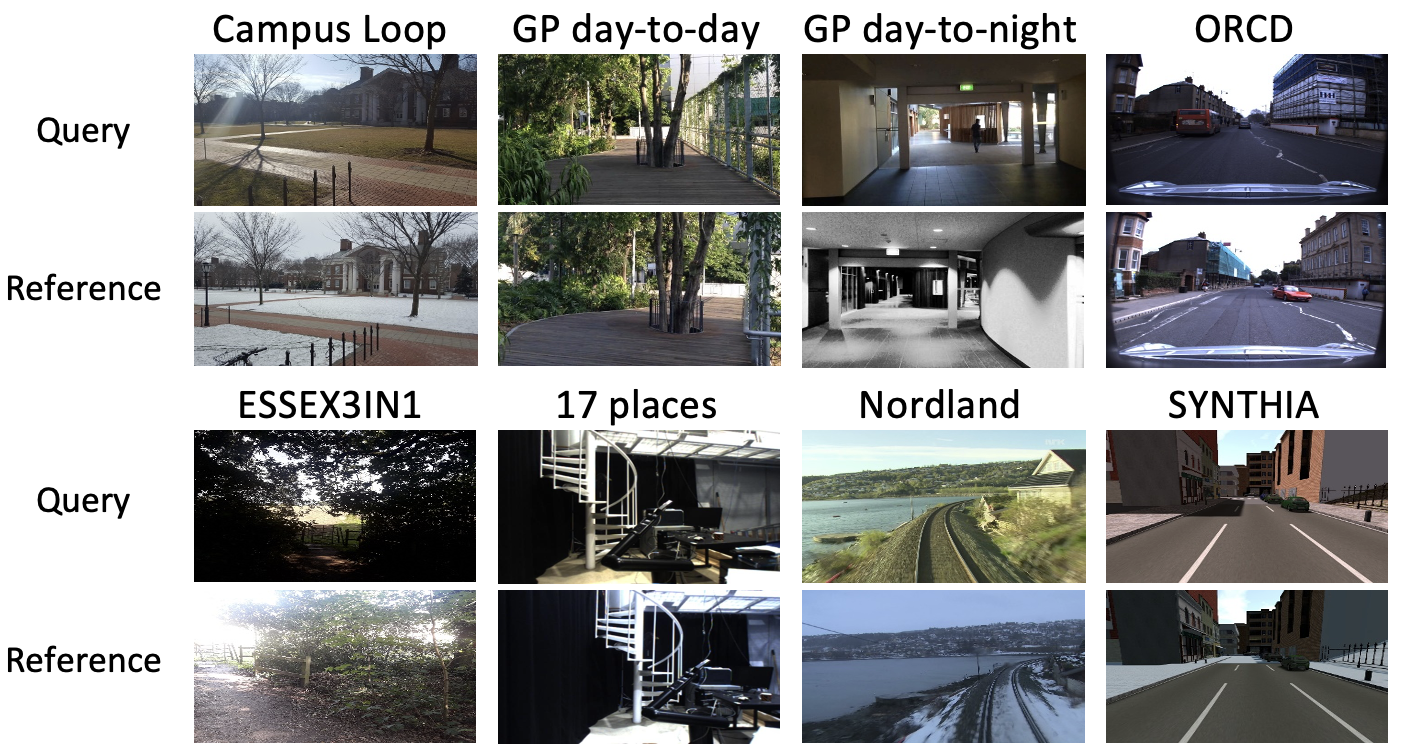}
            \end{tabular}
            \caption{An example of an uncompressed query image and its corresponding reference image taken from each dataset.}
            \label{datasetimagescompression}
        \end{figure}

    On datasets including illumination variation, such as Gardens Point day-to-night and SYNTHIA, techniques such as NetVLAD, CoHOG and HOG lose significant performance throughout the JPEG compression spectrum. The most affected VPR technique on 17 places dataset is CoHOG, where the application of JPEG compression translates to a prominent decrease in performance, as shown in Fig. \ref{VPRperformance}. However, the results for SYNTHIA show that it is slightly more stable than Gardens Point. As SYNTHIA is a synthetic dataset, it is less information rich than a real-word dataset such as Gardens Point (refer to Fig. \ref{datasetimagescompression}). The application of JPEG compression on the SYNTHIA dataset does not alter significantly the image content from the perspective of VPR. This conclusion is supported by Fig. \ref{entropy_compression} that shows the average entropy \cite{zaffar2020cohog}, \cite{tomita2021convsequential} (on the Y-axis) in each query dataset resulting from applying different levels of JPEG compression. The reduction in entropy on SYNTHIA is much smaller when compared to other datasets tested. It is worth mentioning that for both Gardens Point day-to-day and day-to-night datasets we use \textit{day left} images as query images, therefore we only provide the entropy results once in Fig. \ref{entropy_compression}.

        \begin{figure}[t]
            \centering
            \begin{tabular}{ c }
                
                \includegraphics[width=0.95\columnwidth, trim=8 8 8 8, clip]{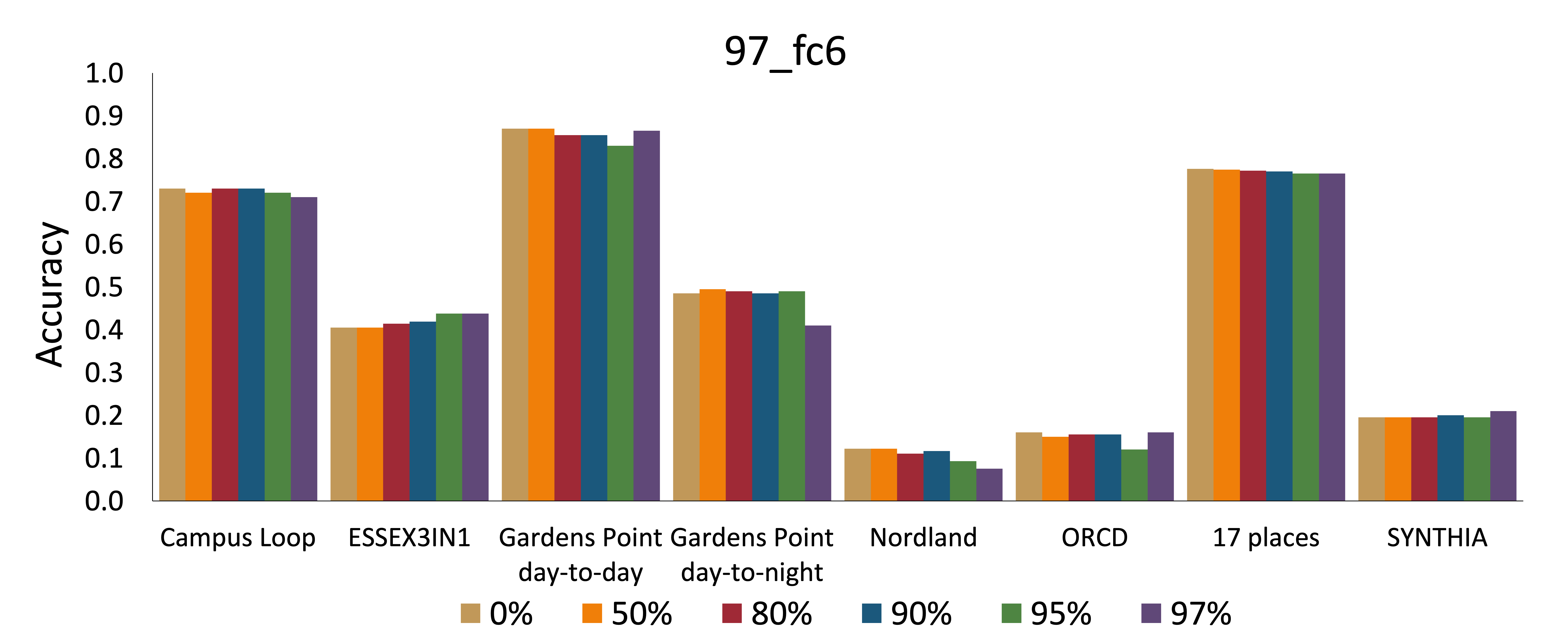}
            \end{tabular}
            \caption{The accuracy of our model on all 8 datasets is enclosed here.}
            \label{accuracy_all_models}
        \end{figure}
    
\subsection{JPEG Optimized CNN} \label{jpgoptimizedcnn}
    \subsubsection{Model Design}\label{modeldesign}
    In an effort to achieve more consistent place matching performance for different levels of JPEG compressed data, we fine-tuned a neural-network based VPR technique specifically for image compression. The neural network has the same structure as AlexNet \cite{AlexNet}, and has been trained on the Places365 dataset \cite{zhou2017places}, which contains approximately 1.8 million images from 365 scene categories. Then, this neural network has been fine-tuned using 97\% JPEG compressed versions of the images taken from the above mentioned dataset. We have specifically selected 97\% JPEG compression rate as it provides the best trade-off between performance and stability.
    The resulting model, entitled \textit{97}, 
    achieves great stability and consistent performance in the higher compression spectrum, on a variety of environments and viewing conditions. Our model achieves a considerable improvement in place matching performance on JPEG compressed data over AlexNet, while at the same time being capable of matching and even outperforming the deeper HybridNet at high compression ratios.
    
    

    \subsubsection{Model Stability and Performance}\label{modelperformance}

    We have previously mentioned in section \ref{utilisedvprtechniques} that each VPR technique has a different performance depending on the type and state of the perceived environment and the quality of the data subtracted from it. To successfully perform VPR tasks in real world applications, it is fundamental to determine the best technique with regards to the above-mentioned environmental variables.  Thus, in this sub-section, we present a comparison between the place matching performance of our model and that of other VPR techniques, throughout the entire spectrum of JPEG compression. 
     
    Fig. \ref{VPRperformance} shows that JPEG compression has drastic effects on the performance of most VPR techniques, especially when using a significant amount of compression (e.g. 97\%). As there is no universal model that achieves the highest VPR performance on all tested datasets, we present in Fig. \ref{accuracy_all_models} the accuracy of one of the best performing models, with different amounts of compression applied to each dataset. We have selected the features from the \textit{fc6} layer as they achieve the best performance on average.

     \begin{figure}[t]
            \centering
            \begin{tabular}{ c }
                
                \includegraphics[width=0.95\columnwidth, trim=8 8 8 8, clip]{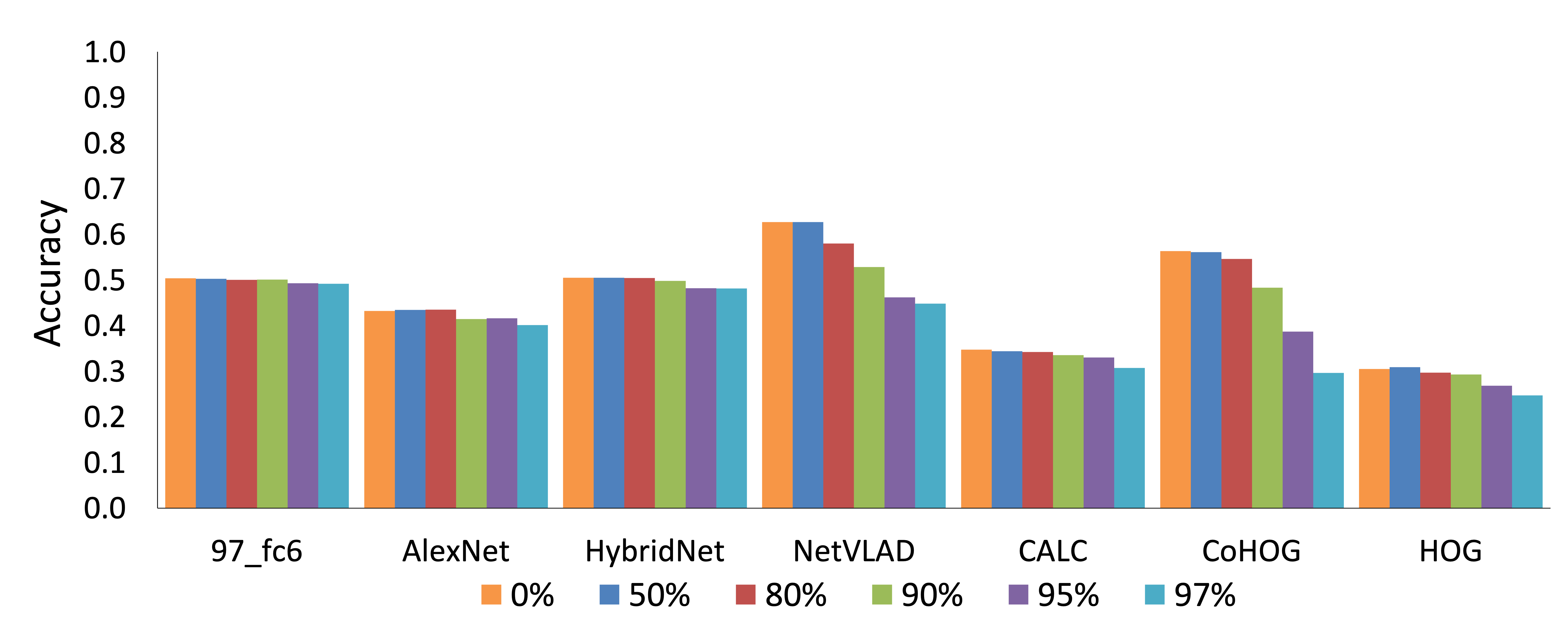}
            \end{tabular}
            \caption{The average accuracy of our model in comparison with other VPR techniques on the combined datasets, for each level of JPEG compression applied.}
            \label{average_performance}
        \end{figure}
    
    The performance of our \textit{97\_fc6} model is shown in Fig. \ref{accuracy_all_models}. The VPR accuracy is highly stable across different amounts of JPEG compression. 
    Fig. \ref{average_performance} compares the average VPR performance between our model and the other VPR techniques, across all tested datasets. The results presented in Fig. \ref{average_performance} show that our model has more consistent performance on compressed data and tends to have a steadier decrease in performance throughout the compression spectrum. As previously mentioned, there is a trade-off in the performance and stability of each VPR technique with respect to the amount of image compression applied. While outperformed on low compression ratios by NetVLAD, HybridNet and CoHOG, our JPEG optimized model can better operate at the highest compression ratios. This is highlighted in Fig. \ref{VPRperformance} on the Campus Loop dataset, where NetVLAD achieves the highest place matching performance for compression levels of up to 90\%. However, when compression becomes extreme (97\%), our model should be used instead as it achieves higher place matching performance as reported in Fig. \ref{accuracy_all_models}. This observation also applies to the Gardens Point day-to-day dataset. Moreover, on Gardens Point day-to-night dataset, our model outperforms every technique for compression levels of above 80\%. Our \textit{97\_fc6} model outperforms AlexNet on all JPEG compressed datasets, except for SYNTHIA as seen in Fig. \ref{VPRperformance}. However, on the 97\% compressed versions of Nordland, Oxford Robot Car and ESSEX3IN1 datasets, our model is outperformed by some VPR techniques presented in Fig. \ref{VPRperformance}. In these cases, the technique that achieves the highest VPR performance should be utilised instead.

\begin{figure}[t]
            \centering
            \begin{tabular}{ c }
                
                \includegraphics[width=0.95\columnwidth, trim=8 8 8 8, clip]{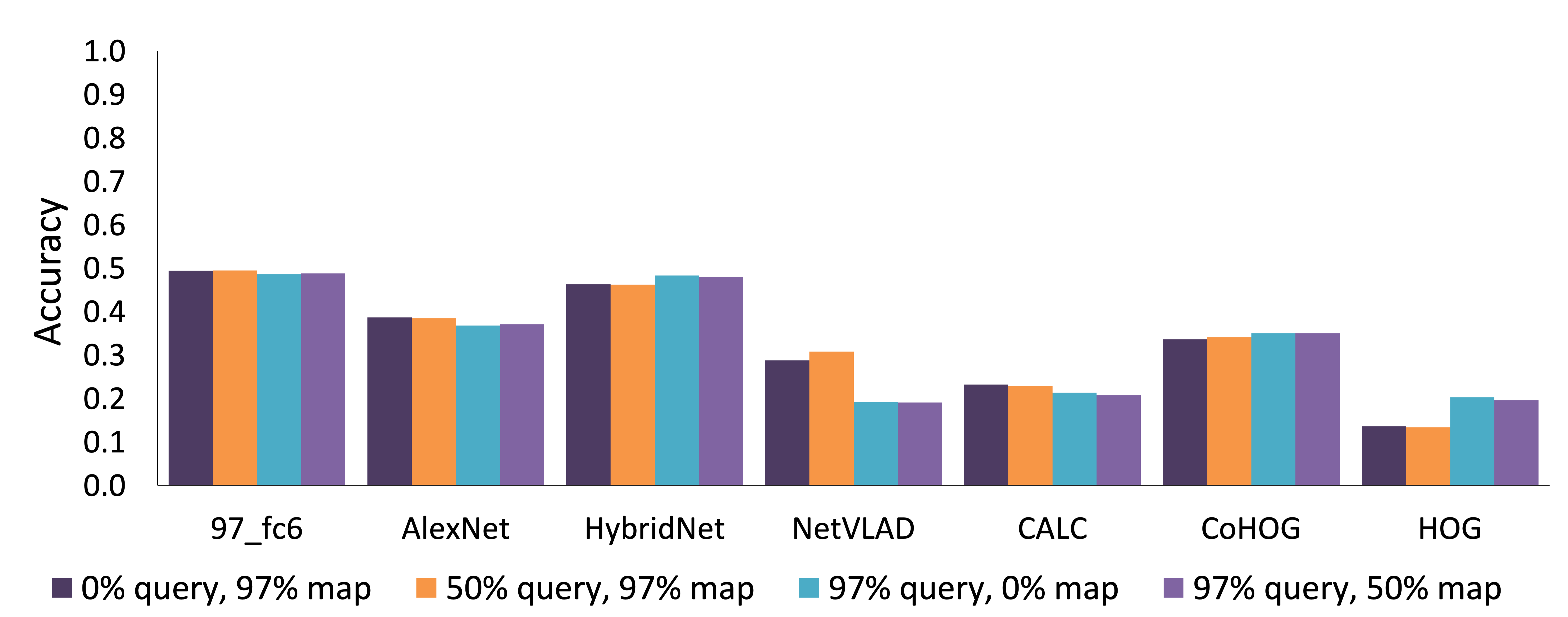}
            \end{tabular}
            \caption{The average place matching performance of our model in comparison with the other VPR techniques presented, in scenarios where the amount of JPEG compression applied to query and reference images greatly differs.}
            \label{nonuniform_compression_average}
        \end{figure}
    
    \begin{figure}[t]
            \centering
            \begin{tabular}{ c c }

                \includegraphics[width=111pt, trim=8 8 8 8, clip]{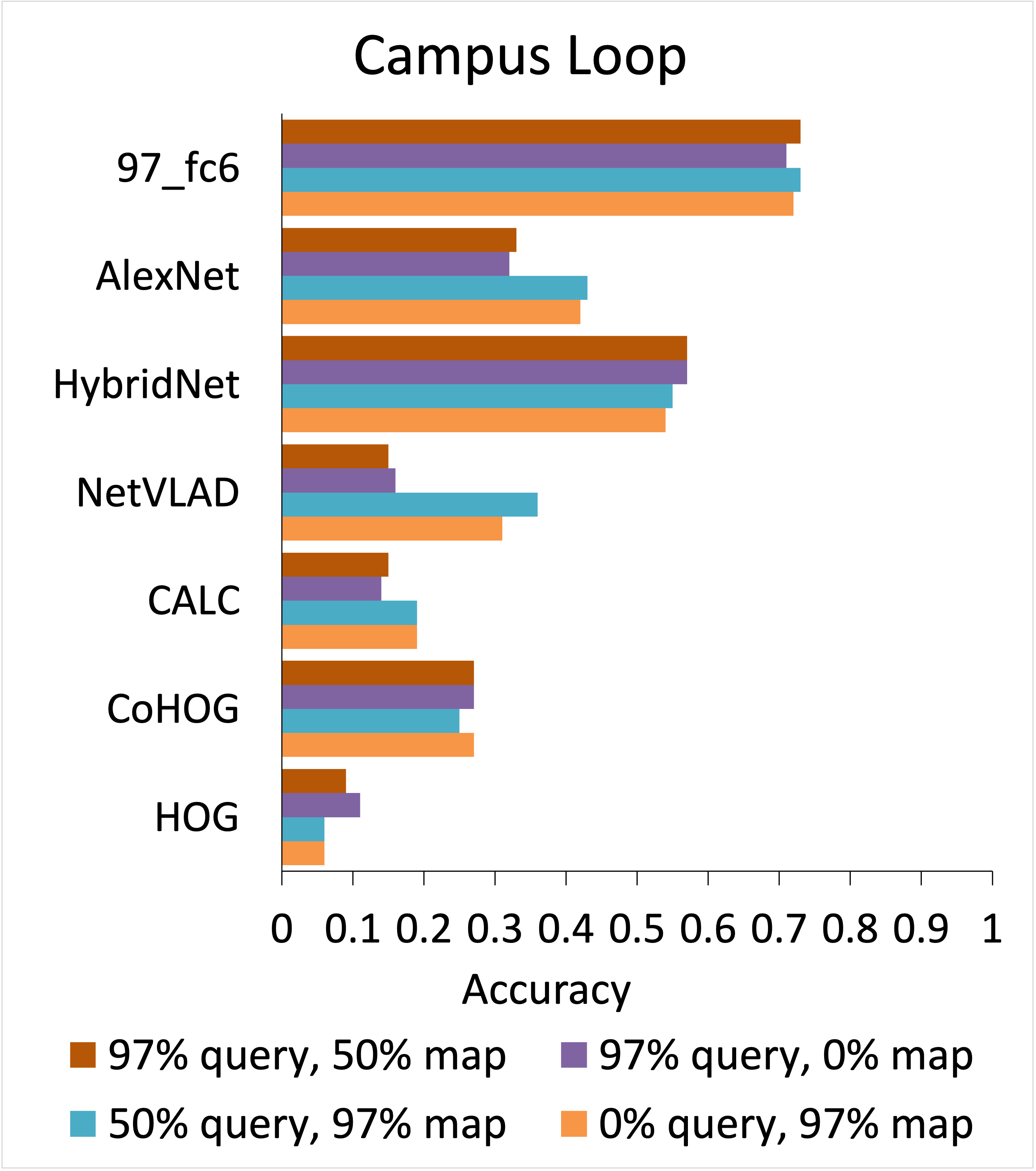} & 
                
                \includegraphics[width=111pt, trim=8 8 8 8, clip]{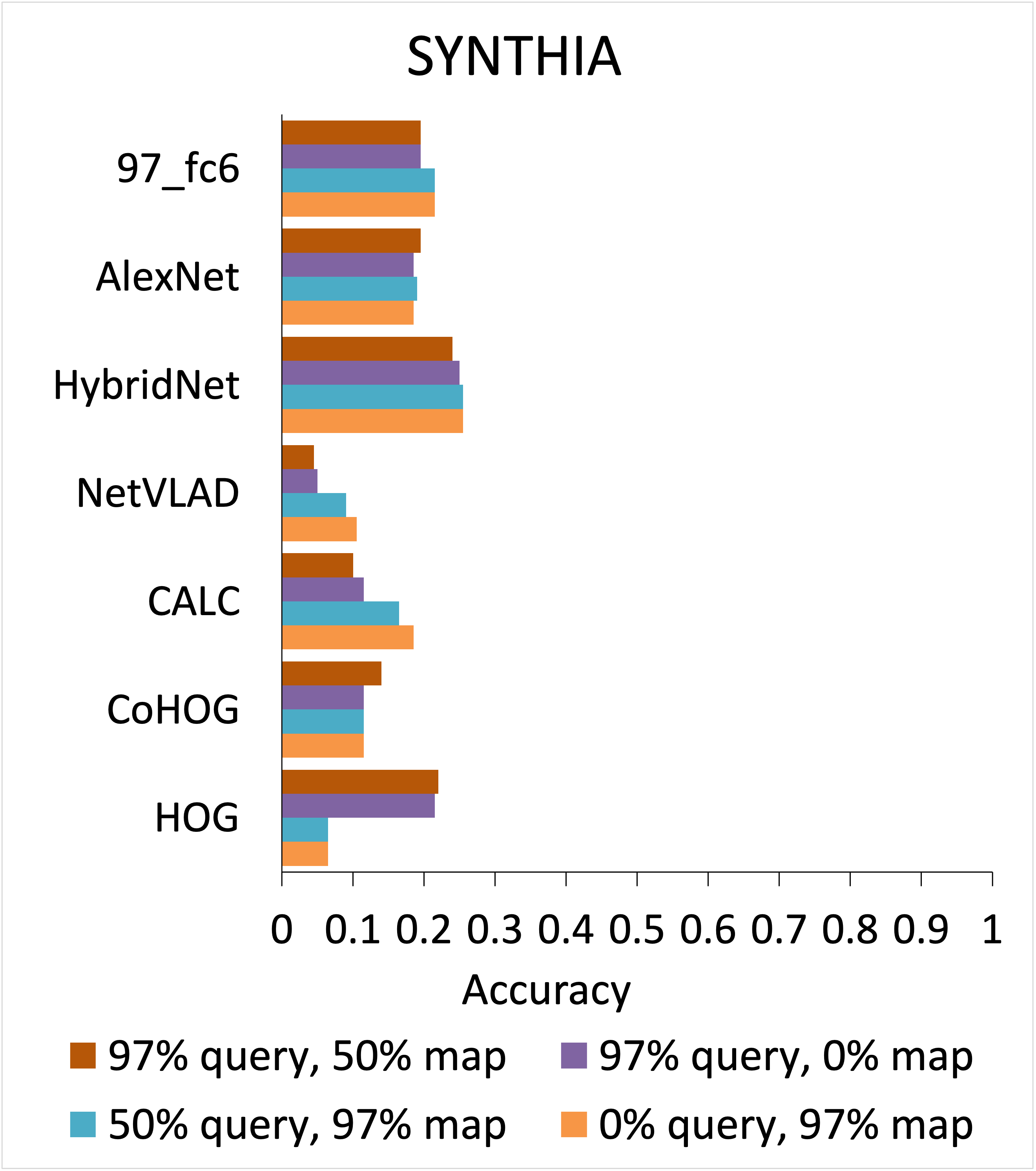} 
                
            \end{tabular}
            \caption{The accuracy of our model in comparison with each VPR technique on the Campus Loop and SYNTHIA datasets, where the query and map images are differently compressed.}
            \label{nonuniform_compression}
            \end{figure}

    \subsection{Non-uniform JPEG Compressed Datasets} \label{nonuniformjpgcompression}
    
    In some practical cases, the visual data transmitted by an agent might be subject to bandwidth limitations, raising the need to use highly compressed images. Hence, the query and stored images (e.g. the map) may have different JPEG compression levels applied (non-uniform compression). 
    For this reason, an analysis on the effects of non-uniform JPEG compression on the performance of our model and other VPR techniques is presented below. 
    
    Fig. \ref{nonuniform_compression_average} presents the average performance of every VPR technique in scenarios where all datasets are non-uniformly compressed. To meet the space constraints, we only included the most significant results, those for the extremes of the JPEG compression spectrum. Apart from being stable on highly JPEG compressed images as discussed in section \ref{modelperformance}, our model also has consistent performance on datasets where the amount of JPEG compression applied to the query and reference images are in the opposite spectrum. Our \textit{97\_fc6} model outperforms AlexNet and achieves slightly better overall performance than HybridNet on non-uniform JPEG compressed data.

    Fig. \ref{nonuniform_compression} presents the detailed results on Campus Loop and SYNTHIA datasets with non-uniform JPEG compression applied. On the Campus Loop dataset, our model achieves the highest VPR performance, outperforming every VPR technique tested. In section \ref{placematchingperformance} we have shown that SYNTHIA is more stable under JPEG compression (due to its synthetic nature) in contrast to other datasets taken from real-world environments. This observation is also emphasized in Fig. \ref{nonuniform_compression}, which shows that the performance of most VPR techniques on the SYNTHIA dataset is not drastically affected in the presence on non-uniform JPEG compression, especially when compared with the results presented in Fig. \ref{VPRperformance}.

    The results presented in Fig. \ref{nonuniform_compression_average} and Fig. \ref{nonuniform_compression} show that our model is more tolerant to non-uniform compression than any of the other VPR techniques tested. Moreover, the results presented throughout this paper emphasize the exceptional performance stability achieved by our neural-network on both uniform and non-uniform JPEG compressed data.

    \section{Conclusions} \label{conclusion}
    This paper conducts an in-depth study on the effects of JPEG compression in VPR. We use a selection of well-established VPR techniques on a variety of JPEG compressed VPR datasets to present our findings. Our experiments show that techniques which are designed to deal with appearance changes present higher tolerance to JPEG compression in comparison with techniques that are designed to handle viewpoint variations. In an attempt to achieve more stable VPR performance when using JPEG compressed data, we demonstrated how fine-tuning can optimise a CNN descriptor to handle highly compressed images. The results show that our model is more consistent on both uniform and non-uniform JPEG compressed data than any other VPR technique presented in this work.
    
    

    {
    \small
    \bibliographystyle{ieeetr}
    \bibliography{root}
    }
    
\end{document}